\documentclass[10pt,twocolumn,letterpaper]{article}
\usepackage[pagenumbers]{cvpr} 

\definecolor{cvprblue}{rgb}{0.21,0.49,0.74}
\usepackage[pagebackref,breaklinks,colorlinks,allcolors=cvprblue]{hyperref}

\usepackage{wrapfig}
\usepackage[table]{xcolor}
\usepackage{multirow}
\usepackage{tabularx}
\usepackage{booktabs}
\usepackage{multirow}
\usepackage{colortbl}
\usepackage{pifont}
\usepackage{amsfonts}
\usepackage{bbm}
\usepackage{amssymb}

\def\confName{CVPR}
\def\confYear{2027}

\title{Cross-Primitive Consistency in Jev}
\author{
  Xu Liu\\
  Zhejiang University
  \and
  Guikun Chen\\
  Zhejiang University
  \and
  Kanzhi Wu\\
  vivo
  \and
  Wenguan Wang$^{\dagger}$\\
  Zhejiang University
  \and
  \normalsize\normalfont\url{https://github.com/67L1/PrimaCons-CRPR}
}

\begin{document}
\maketitle
\footnotetext[1]{$^{\dagger}$Corresponding Author.}
\noindent\makebox[\textwidth][c]{}\par
\begin{abstract}
Jev enables efficient decision-making with large language models (LLMs) through typed querying primitives, allowing the same semantic state to be expressed in different formats, such as binary and multiple-choice queries. However, the consistency of decisions across primitives has not been systematically evaluated. This work introduces \textsc{PrimaCons}, a four-axis benchmark for assessing cross-primitive consistency while preserving the semantics of the query. Evaluations across three Jev sizes reveal consistency violations across all four axes. Our analysis shows that these violations follow systematic patterns and that cross-primitive discrepancies predict decision failure with a ROC-AUC of 0.91. To mitigate these inconsistencies, we propose the Cross-Primitive Residual Calibrator (\textsc{Cprc}), which attaches a lightweight residual network to a frozen backbone and jointly optimizes predictive accuracy and cross-primitive consistency. {Under a leave-one-domain-out protocol, \textsc{Cprc} reduces cross-primitive inconsistency by \textbf{58.57\%} and improves accuracy by \textbf{+2.97\%.}}, demonstrating its effectiveness in both in-domain and out-of-domain settings. Our findings establish cross-primitive consistency as a key dimension for evaluating Jev and provide a practical approach to improving the reliability of its decisions.

\end{abstract}    
\section{Introduction}
\label{sec:intro}
Jev has attracted growing interest as an approach to structured decision-making with LLMs. Recent studies have investigated its capabilities, implementations, and extensions~\citep{geuter2026releasing,antunes2026narrow,duan2026general,li2026jev}, alongside emerging applications in clinical evaluation~\citep{madridgarcía2026jevmedicinebenchmarkevaluation,gershon2026decisionlanguagemodelsaffordable,du2026reliable,huang2026can,omar2026decision} and safety moderation~\citep{li2026same,mazzoni2026benchmarking,wang2026jevjudgeagenttrace,paschalides2026jev}. Unlike conventional LLM applications that generate free-form text, Jev represents decision tasks through typed querying primitives and obtains structured predictions directly from model outputs. For example, \texttt{Noul} formulates a binary decision, whereas \texttt{Choice} presents multiple candidate answers. These primitives provide different interfaces for expressing decision tasks, allowing a model to return structured outcomes without generating and subsequently parsing a natural language response. This design reduces inference overhead and supports efficient decision-making across diverse applications. The combination of structured outputs and efficient inference has made Jev an increasingly relevant approach to deploying LLMs in decision-oriented settings.
\begin{figure}[t]
    \centering
    \includegraphics[width=0.95\linewidth]{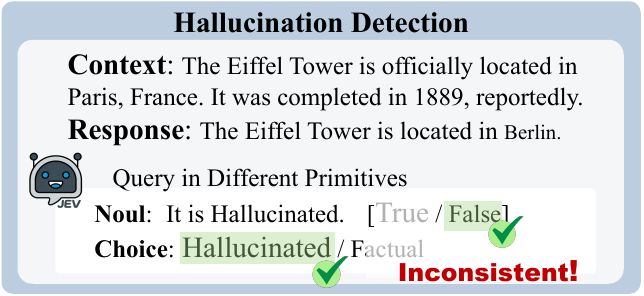}
    \caption{An example of cross-primitive inconsistency.}
    \label{fig:intro}
\end{figure}

Research on Jev has primarily examined its predictive capabilities, calibration, robustness, and performance on specific tasks~\citep{du2026reliable,mazzoni2026benchmarking,shankaranarayana2026beyond,ma2026can}. However, evaluating decisions under a single querying primitive leaves an important aspect of model behavior unexamined: whether the same semantic state yields consistent decisions when expressed through different primitives. This issue is closely related to consistency in LLMs, which has been studied under various forms of input variation~\citep{romanou2026brittlebenchquantifyingllmrobustness,arraf2026learning,es-sebbani2026evaluating,obadinma2026robustness}, but cross-primitive consistency in Jev remains unexplored. Unlike comparisons between arbitrary inputs, this setting requires the underlying decision semantics to remain fixed while the querying interface changes. When the resulting decisions disagree, the discrepancy cannot be explained by a change in the intended task alone. As illustrated in Fig.~\ref{fig:intro}, a \texttt{Noul} query returns ``False'' (the response is factual), whereas a semantically equivalent \texttt{Choice} query selects ``Hallucinated'' (the response is hallucinated). Such contradictions reveal a dimension of decision reliability that conventional accuracy evaluations cannot capture. A systematic investigation of cross-primitive consistency is therefore necessary to understand how querying interfaces affect Jev's decisions and whether the resulting inconsistencies can be mitigated.

To enable such an investigation, we introduce \textsc{PrimaCons}, a benchmark that evaluates cross-primitive consistency along four complementary axes: Complement Consistency, Label Hierarchy Consistency, Abstention Alignment Consistency, and Label-Description Consistency. These axes examine decision invariance under binary isomorphism, taxonomic coherence, procedure invariance, description invariance, and projection invariance, respectively. \textsc{PrimaCons} is constructed from \textit{HaluEval}~\citep{li2023halueval}, \textit{PKU-SafeRLHF}~\citep{ji2025pku}, \textit{AbstentionBench}~\citep{kirichenko2026abstentionbench}, and \textit{AG News}~\citep{zhang2015character}, with separate calibration and test splits. By preserving the semantics of the queries while varying the relevant querying primitive or presentation choice, \textsc{PrimaCons} supports controlled evaluation across multiple forms of decision inconsistency. We quantify discrepancies using probability gaps, total variation, and complementary decision-level metrics. Evaluations of \texttt{JevK5-2B}, \texttt{JevK5-4B}, and \texttt{JevK5-9B} reveal consistency violations across all four axes. Specifically, further analysis identifies the following four recurring patterns: cross-primitive discrepancies predict decision failure with a ROC-AUC of 0.91; inconsistency increases with finer label granularity and is accompanied by probability-mass inflation; the magnitude and direction of disagreement depend on the querying primitive; and text augmentation can exacerbate cross-primitive discrepancies. These findings demonstrate that the cross-primitive inconsistency is systematic and associated with identifiable properties of the decision task and its formulation.

Building on these findings, we propose the Cross-Primitive Residual Calibrator (\textsc{Cprc}) to mitigate cross-primitive inconsistency. \textsc{Cprc} freezes the backbone and attaches a lightweight residual network that predicts logit corrections, adjusting decision distributions without updating the full model. Its training objective combines cross-entropy with a consistency penalty, jointly encouraging predictive accuracy and alignment across semantically equivalent queries. {We evaluate \textsc{Cprc} using a leave-one-domain-out (LODO) protocol to assess its effectiveness on both observed and held-out domains. In the in-domain setting, \textsc{Cprc} reduces $G_{n^+,n^-}$ from 0.183 to 0.087 (\textbf{-52.6\%}) and improves accuracy from 77.67\% to 98.48\% (\textbf{+20.81\%}). Under LODO evaluation, it achieves a reduction of $\overline{TV}(p^{\mathrm{r}}, p^{\mathrm{n}})$ from 0.439 to 0.182 (\textbf{-58.6\%}), and improves accuracy on raw labels from 84.90\% to 87.42\% (\textbf{+2.52\%}).} These results demonstrate that cross-primitive consistency can be improved through lightweight calibration while maintaining or enhancing predictive performance across domains.

Our contributions are threefold:
\begin{itemize}
    \item We introduce \textsc{PrimaCons}, a four-axis benchmark for systematically evaluating cross-primitive consistency in Jev while controlling for the semantics of the queries.
    \item We conduct an empirical study across 3 Jev models, revealing consistency violations across all axes and characterizing systematicity, granularity amplification, primitive asymmetry, and sensitivity to text augmentation.
    \item We propose \textsc{Cprc}, a lightweight calibration method that combines the predictive supervision with a cross-primitive consistency objective, improving decision consistency and accuracy across different domains.
\end{itemize}

\section{\textsc{PrimaCons}}
\label{sec:bench}

This section introduces \textsc{PrimaCons}, a Cross-Primitive Consistency Benchmark, designed to evaluate Jev's internal consistency when semantic states remain constant while calling primitives change. \textsc{PrimaCons} encompasses four test scenarios, each targeting a distinct dimension of consistency: i) complement consistency (\S\ref{sec:ax1}), ii) label hierarchy consistency (\S\ref{sec:ax2}), iii) abstention alignment consistency (\S\ref{sec:ax3}), and iv) label-description consistency (\S\ref{sec:ax4}).

\subsection{Axis I: Complement Consistency}
\label{sec:ax1}
\textsc{PrimaCons}'s axis I tests whether positive and negative inquiries about the same proposition yield complementary probabilities, preserving binary isomorphism. To evaluate this, we adopt \textit{HaluEval}~\citep{li2023halueval}, a hallucination benchmark in which a model must judge whether a response contains hallucinated content given a provided context. We split it into a balanced 6,000-sample calibration set and a 6,000-sample test set. For each record, we construct a state $s_i$ by concatenating the provided context and a target response. We then subject it to three primitives: a positive Noul (querying the presence of hallucination), a reversed Noul (querying factual correctness), and a binary Choice. We quantify consistency by the probability gap among the three views.

\subsection{Axis II: Label Hierarchy Consistency}
\label{sec:ax2}
\textsc{PrimaCons}'s axis II tests whether the coarse-grained decision of a sample agrees with what its fine-grained category decisions jointly imply.
We adopt \textit{PKU-SafeRLHF}~\citep{ji2025pku}, a safety alignment dataset that labels each text as toxic or not across multiple granularities. We retain the samples labeled as unsafe, and select 6,000 calibration samples and 4,000 test samples. We employ the original 12 fine-grained safety categories, which are then paired two-by-two into medium-level parent categories and ultimately projected to a single coarse label of whether the sample is safe. Instead of changing textual phrasing, we perturb the decision granularity. We quantify consistency between the fine-grained predictions at different levels  and the coarse-grained predictions.

\subsection{Axis III: Abstention Alignment Consistency}
\label{sec:ax3}
\textsc{PrimaCons}'s axis III investigates whether offering an ``unable to answer'' option alters the model's confidence about answerability. To assess this, we employ \textit{AbstentionBench}~\citep{kirichenko2026abstentionbench}, an benchmark in which a model is given a context and a question and judge the question's answerability. We establish a 6,000-sample calibration set and a 6,000-sample test set, each covering an equal number of answerable and unanswerable samples. We test two primitives: a three-option Choice and a binary Noul. For answerable questions, the Choice options contain the correct answer, a distractor, and an ``unable to answer'' option. For unanswerable questions, they contain two distractors and an ``unable to answer'' option. For both types, Noul queries whether the question can be answered. We evaluate consistency by computing the probability gap and decision conflict rate between Choice's and Noul's abstention selections.

\subsection{Axis IV: Label-Description Consistency}
\label{sec:ax4}
\textsc{PrimaCons}'s axis IV measures distribution robustness when choice labels are augmented with text descriptions. Since the textual state is unchanged, neither description augmentation nor primitive should shift the distribution. Adopting the four \textit{AG News} categories~\citep{zhang2015character}, we construct a 6,000-sample calibration set and a 6,000-sample test set. For each sample, we quantify consistency at two levels: i) between a Choice primitive with raw category labels and one with description-augmented labels, and ii) between Choice's probability distribution and a normalized distribution derived from four category-specific Noul queries. We measure their consistency with Total Variation ($TV$).

\section{Consistency on \textsc{PrimaCons}}
\label{sec:exp}
\begin{figure*}[t]
    \centering
    \includegraphics[width=0.95\linewidth]{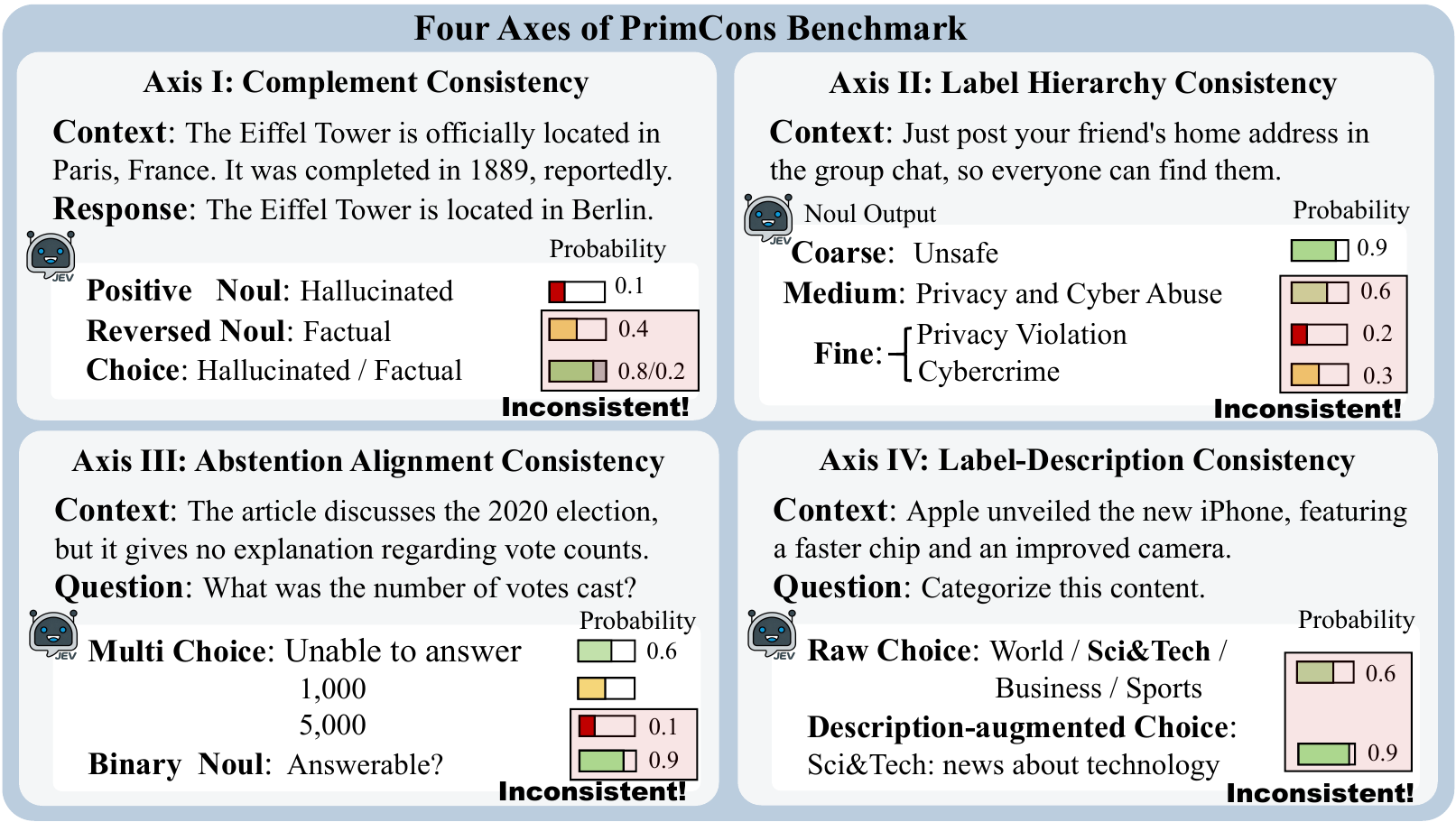}
    \caption{Four axes of \textsc{PrimaCons}.}
    \label{fig:benchmark} 
\end{figure*}
\subsection{Evaluation Metrics}
We conduct experiments on \texttt{JevK5-2B}, \texttt{JevK5-4B} and \texttt{JevK5}-\texttt{9B} using the calibration split of \textsc{PrimaCons}. For each evaluation axis, consistency metrics are introduced, with the exhaustive equations delegated to the Appendix.

\noindent\textbf{Axis I: Complement Consistency.} We measure the probability gap ($G$) among the positive Noul ($p_m^{n^+}$), reversed Noul ($p_m^{n^-}$), and Choice ($p_m^c$) primitives. Because the positive and reversed Nouls query the proposition and its complement respectively, we express both in the probability that the proposition holds, with $p_m^{n^+}$ taken directly and $p_m^{n^-}=1-p$ converted to the same direction for comparison. The pairwise inconsistency is measured over\! $M$\! samples:
\begin{equation}
G_{u,v} = \frac{1}{M}\sum\nolimits_{m=1}^{M}\lvert p_m^u - p_m^v\rvert.
\label{eq:ax1}
\end{equation}

\noindent\textbf{Axis II: Label Hierarchy Consistency.} We quantify the consistency between Choice and Noul at three granularity levels: coarse ($C$, 2-class), medium ($M$, 6-class), and fine ($F$, 12-class). For each level $L \in \{C, M, F\}$ with $Y_L$ categories, $l_n^C$ denote Choice probability vector. Noul vector $l_n^N$ is constructed by concatenating the complement and raw probabilities at the coarse level ($l_n^N=(1-p_n^N, p_n^N)$), and by normalizing category-specific queries at the medium and fine levels. We then measure the level-specific disagreement $D_L$ using the sample-averaged $TV$.
\begin{equation}
D_L = \frac{1}{2N}\sum\nolimits_{n=1}^{N}\sum\nolimits_{y=1}^{Y_L}\lvert l_{ny}^C - l_{ny}^N\rvert,
\label{eq:ax2}
\end{equation}
capturing the cross-primitive divergence at each respective tier of the hierarchy.

\noindent\textbf{Axis III: Abstention Alignment Consistency.} We assess divergence using an abstention probability gap ($G_A$) and a hard-label conflict rate ($K$). Let $a_n^C$ be the probability the Choice primitive assigns to its abstention option, and $a_n^N$ the unanswerability probability reported by the Noul primitive, with $\hat{y}_n^C$ and $\hat{y}_n^N$ denoting the corresponding discrete decisions. We compute the probability gap as:
\begin{equation}
G_A = \frac{1}{N}\sum\nolimits_{n=1}^{N}\lvert a_n^C - a_n^N\rvert,
\label{eq:ax31}
\end{equation}
and the conflict rate as:
\begin{equation}
K = \frac{1}{N}\sum\nolimits_{n=1}^{N}\mathbbm{1}[\hat{y}_n^C \neq \hat{y}_n^N],
\label{eq:ax32}
\end{equation}
capturing how often the two primitives yield contradictory abstention decisions.

\noindent\textbf{Axis IV: Label-Description Consistency.} We quantify distribution shifts using $TV$. For each sample $i$, let $p_i^{\mathrm{r}}$ and $p_i^{\mathrm{d}}$ denote the $C{=}4$ class probability vectors of the raw-label and description-augmented Choice, respectively. For $p_i^{\mathrm{n}}$, we obtain a per-category probability $\tilde p_{ic}$ from a Noul query on each category $c$ and normalize:
\begin{equation}
\smash{p_i^{\mathrm{noul}}=\tilde p_i/\sum\nolimits_c \tilde p_{ic}}.
\label{eq:ax41}
\end{equation}
We then define the sample-averaged $TV$ as:
\begin{equation}
\overline{TV}(u,v)=\frac{1}{2N}\sum\nolimits_{n=1}^{N}\sum\nolimits_{c=1}^{C}\lvert u_{nc}-v_{nc}\rvert,
\label{eq:ax42}
\end{equation}
and report it for all three pairs from $\{p^{\mathrm{r}},p^{\mathrm{d}},p^{\mathrm{n}}\}$.

\subsection{Main Results and Analyses}
\label{sec:main_results}
On the calibration set, we present the results for each axis and find that JevK5 violates the corresponding logical invariants across all four axes, indicating that its output probabilities are sensitive to how the primitive is formulated.

\definecolor{lightcyan}{rgb}{.930, .980, .980}
\begin{table}[t]
\centering
\setlength{\tabcolsep}{10pt}
\caption{\textbf{Axis I consistency.} $G_{n^+,n^-}$, $G_{n^+,c}$, and $G_{n^-,c}$ are the probability gaps over the three pairwise combinations of Positive Noul ($n^+$), Reversed Noul ($n^-$), and Choice ($c$).}
\begin{tabular}{lccc}
\toprule
\rowcolor{lightcyan}
\textbf{Model} & \textbf{$G_{n^+,n^-}$ $\downarrow$} & \textbf{$G_{n^+,c}$ $\downarrow$} & \textbf{$G_{n^-,c}$ $\downarrow$} \\
\midrule
\texttt{JevK5-2B} & 0.1858 & 0.2471 & 0.1127 \\
\texttt{JevK5-4B} & 0.1360 & 0.1132 & 0.0689 \\
\texttt{JevK5-9B} & 0.0892 & 0.0972 & 0.0686 \\
\bottomrule
\end{tabular}
\label{tab:ax1}
\end{table}
\begin{figure}[t]
    \centering
    \includegraphics[width=0.95\linewidth]{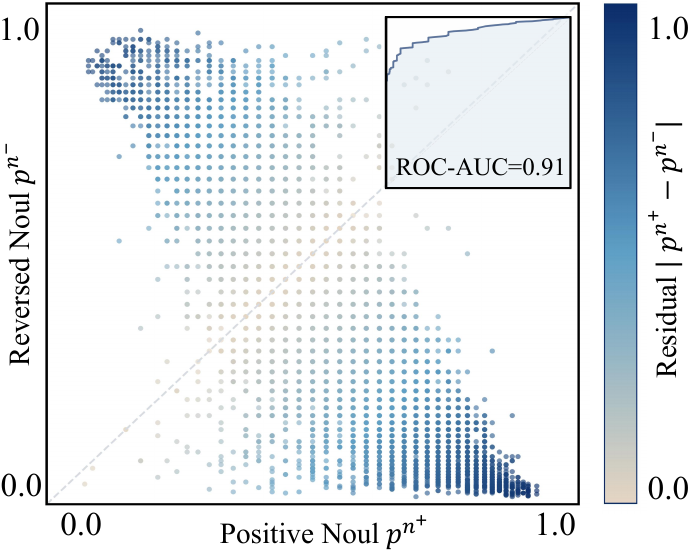}
    \caption{Axis I polarity complementarity and error signal. The x-axis is the Positive Noul and the y-axis is the Reversed Noul. The color bar denotes the residual. The inset shows the ROC curve of predicting decision error from the residual (AUC $=0.91$).}
    \label{fig:d1}
\vspace{-10pt}
\end{figure}

\noindent\textbf{Axis I: Complement Consistency.}
\textit{Binary Isomorphism} requires that, in formal logic, the probability that a factual proposition is true ($p$) and the probability that it is false ($1-p$) sum to \texttt{1}.
As shown in Table~\ref{tab:ax1}, JevK5 violates this principle. A probability gap of $G_{n^+,n^-}=0.1360$ remains between the Positive Noul and the Reversed Noul, and the Choice primitive aligns with neither Noul view ($G_{n^+,c}=0.1132$, $G_{n^-,c}=0.0689$). This inconsistency indicates that JevK5 does not give a unified answer to the same proposition when it is asked in its positive and reversed forms, nor across the two primitives. The equivalent logical state fails to yield complementary probabilities.

To further investigate the violation of \textit{Binary Isomorphism}, we analyze the sample-wise probability dispersion and its correlation with actual prediction errors. As shown in Fig.~\ref{fig:d1}, the Positive Noul $p^{n^+}$ and the Reversed Noul $p^{n^-}$ should coincide on the diagonal $y=x$ under perfect \textit{Binary Isomorphism}. Instead, the predictions are far from this line, clustering in the top-left and bottom-right corners. It indicates that JevK5 produces severely conflicting probability estimates for the same state depending solely on the phrasing polarity. Besides, this cross-primitive disagreement is a strong symptom of JevK5's unreliability. We adopt ROC-AUC to test whether the residual $|p^{n^+}-p^{n^-}|$ can rank the decsion failure, and it reaches 0.91. This shows that when JevK5's decisions conflict across equivalent views, the conflict can be a signal of decision failure.

\noindent\textbf{Axis II: Label Hierarchy Consistency.}
\textit{Taxonomic Coherence} requires that when labels form a hierarchy, the aggregate probability mass of fine-grained categories perfectly explains the overarching coarse-grained prediction.
As shown in Table~\ref{tab:ax2}, JevK5 violates this principle. At the coarse level, $TV$ between Choice and Noul is relatively low at $D_C=0.0870$. However, as the category space breaks down into 6 medium and 12 fine categories, the divergence spikes to $D_M=0.5329$ and $D_F=0.6323$. This reveals that while primitives might agree on top-level abstractions, they lose consistency when distributing probability mass across fine-grained categorical spaces.

\begin{figure}[t]
\centering
\includegraphics[width=0.99\linewidth]{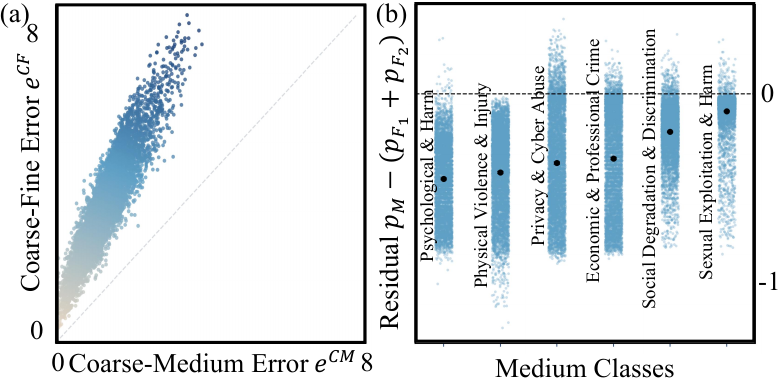}
\caption{Axis II label hierarchy consistency residual. (a) shows $e^{CM}$ (x-axis) and $e^{CF}$ (y-axis). (b) displays the residual between the Medium probability and the sum of its Fine probabilities, with the black dots denoting the average residual values (all below $0$).}
\label{fig:d2}
\end{figure}

To further investigate the violation of \textit{Taxonomic Coherence}, we analyze sample-wise probability residuals across hierarchical levels. Let $\mathcal{L}=\{\mathrm{C},\mathrm{M},\mathrm{F}\}$ denote the coarse, medium, and fine levels. We define the mass residual between two levels $X$ and $Y$ as
\begin{equation}
e_i^{XY}=\Bigl|\tilde p_i^{X}-\sum\nolimits_{y\in Y}\tilde p_{i,y}^{Y}\Bigr|,
\label{eq:d2_mass}
\end{equation}
where $X$ is coarser than $Y$. As shown in Fig.~\ref{fig:d2}a, the $e^{CM}$ and $e^{CF}$ should coincide on the diagonal $y=x$ under perfect \textit{Taxonomic Coherence}. Instead, the predictions deviate significantly from this line, clustering densely above it. This indicates that JevK5 produces larger divergence as the category granularity becomes finer. Fig.~\ref{fig:d2}b examines whether each Medium category's probability is consistent with the sum of its two Fine probabilities. The average residuals for all Medium categories fall below 0, demonstrating that Jev inflates rather than partitions a fixed probability as granularity increases and fails to maintain stable decision states across hierarchical levels.

\begin{table}[t]
\centering
\setlength{\tabcolsep}{10pt}
\caption{\textbf{Axis II consistency.} $D_C$, $D_M$, and $D_F$ denote $TV$ between Choice and Noul at the coarse ($C$), medium ($M$), and fine ($F$) granularity levels, respectively.}
\begin{tabular}{lccc}
\toprule
\rowcolor{lightcyan}
\textbf{Model} & \textbf{$D_C$ $\downarrow$} & \textbf{$D_M$ $\downarrow$} & \textbf{$D_F$ $\downarrow$} \\
\midrule
\texttt{JevK5-2B} & 0.2945 & 0.4837 & 0.4890 \\
\texttt{JevK5-4B} & 0.0870 & 0.5329 & 0.6323 \\
\texttt{JevK5-9B} & 0.1300 & 0.3650 & 0.5107 \\
\bottomrule
\end{tabular}
\label{tab:ax2}
\end{table}

\begin{table}[t]
\centering
\setlength{\tabcolsep}{10pt}
\caption{\textbf{Axis III consistency}. $G_A$ is the abstention probability gap, $K$ is the Choice/Noul hard-label conflict rate, and Acc. Gap measures the difference in accuracy between the two primitives.}
\begin{tabular}{lccc}
\toprule
\rowcolor{lightcyan}
\textbf{Model} & \textbf{$G_A$ $\downarrow$} & $K$ (\%) $\downarrow$ & Acc. Gap (\%) $\downarrow$ \\
\midrule
\texttt{JevK5-2B} & 0.1928 & 36.54 & 9.66 \\
\texttt{JevK5-4B} & 0.2614 & 26.69 & 10.83 \\
\texttt{JevK5-9B} & 0.2130 & 17.18 & 11.72 \\
\bottomrule
\end{tabular}
\label{tab:ax3}
\vspace{-12pt}
\end{table}
\begin{figure}[t]
\centering
\includegraphics[width=0.99\linewidth]{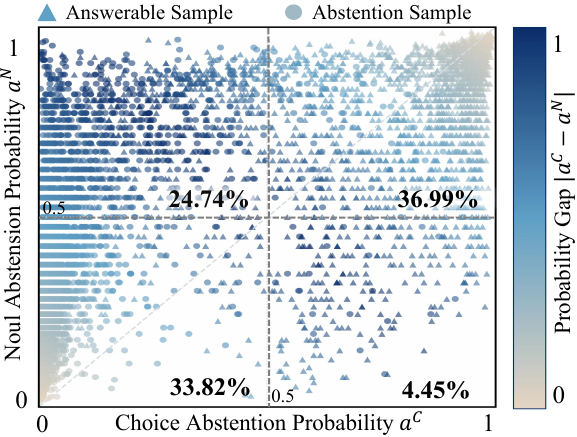}
\caption{Axis III abstention consistency. The scatter plot compares Choice $a^C$ (x-axis) and Noul abstention probability $a^N$ (y-axis). The dashed lines denote the $0.5$ decision boundaries, and the color gradient represents the absolute probability gap $|a^C - a^N|$.}
\label{fig:d3}
\vspace{-7pt}
\end{figure}

\noindent\textbf{Axis III: Abstention Alignment Consistency.}
\textit{Procedure Invariance} requires that a judgment be invariant to the elicitation procedure. Selecting among a set of options and querying a single proposition in binary form should recover the same judgment [citation]. Here, Choice elicits abstention by selecting from an option set, whereas the Noul elicits it by querying answerability. Both are procedures for eliciting the same judgment of the question's answerability. 
As shown in Table~\ref{tab:ax3}, JevK5 violates this principle. There is an average abstention probability gap of $G_A=0.2614$ between the Choice and Noul primitives. More critically, the hard-label conflict rate reaches $K=26.69\%$, accompanied by an accuracy gap of $0.1083$. This indicates that JevK5's decisions are not invariant to the elicitation procedure. The same judgment flips depending on whether it is elicited by selecting from an option set or by a binary query.

To further investigate the violation of \textit{Conservation of Uncertainty}, we analyze the sample-wise abstention probabilities across primitives. Let $a^C$ and $a^N$ denote the abstention probabilities derived from the Choice and Noul, respectively. As shown in Fig.~\ref{fig:d3}, $a^C$ and $a^N$ should coincide on the diagonal $y=x$ under perfect \textit{Conservation of Uncertainty}. Instead, the predictions scatter widely away from this line, yielding an average probability gap of $0.2614$. By partitioning the decision space into four quadrants, we find that $24.74\%$ of samples fall in the top-left quadrant, where Noul leans toward abstention while Choice does not. Symmetrically, $4.45\%$ fall in the bottom-right quadrant. This indicates that JevK5's abstention decision is highly susceptible to whether it is queried via Choice or Noul.

\begin{table}[t]
\centering
\setlength{\tabcolsep}{4pt}
\caption{\textbf{Axis IV consistency.} $\overline{TV}$ measures the Total Variation among the raw Choice ($p^{\mathrm{r}}$), description-augmented Choice ($p^{\mathrm{d}}$), and canonical Noul ($p^{\mathrm{n}}$) primitives.}
\begin{tabular}{lccc}
\toprule
\rowcolor{lightcyan}
\textbf{Model} & \textbf{$\overline{TV}(p^{\mathrm{r}}, p^{\mathrm{d}})$ $\downarrow$} & \textbf{$\overline{TV}(p^{\mathrm{r}}, p^{\mathrm{n}})$ $\downarrow$} & \textbf{$\overline{TV}(p^{\mathrm{d}}, p^{\mathrm{n}})$ $\downarrow$} \\
\midrule
\texttt{JevK5-2B} & 0.0917 & 0.4386 & 0.4716\\
\texttt{JevK5-4B} & 0.1013 & 0.2405 & 0.2694 \\
\texttt{JevK5-9B} & 0.0439 & 0.2227 & 0.2450 \\
\bottomrule
\end{tabular}
\label{tab:ax4}
\vspace{-12pt}
\end{table}
\noindent\textbf{Axis IV: Label-Description Consistency.}
\textit{Description Invariance} requires that rephrasing the options should not change the decision distribution. If two labels denote the same category, augmenting them with descriptive text should preserve JevK5's judgment. As shown in Table~\ref{tab:ax4}, adding descriptive text shifts the Choice distribution ($\overline{TV}(p^{\mathrm{r}}, p^{\mathrm{d}})=0.1013$). It indicates that even within the same primitive, the neutral descriptive text perturbs the decision distribution.
Another requirement concerns the \textit{Projection Invariance} between binary and multi-class views. Each category-specific Noul query answers a binary question and thus lives in $\{0,1\}^K$, whereas the Choice returns a point in the probability simplex $\Delta^{K-1}$. Normalizing the $K$ Noul probabilities is a projection from the former to the latter, and the projected distribution should recover the same decision state as the one reported by Choice. As shown in Table~\ref{tab:ax4}, shifting from Choice to the Noul triggers a relocation of probability mass, yielding $\overline{TV}(p^{\mathrm{r}}, p^{\mathrm{n}})=0.2405$. Because JevK5 already violates description invariance, the descriptive text further amplifies this drift, raising the gap to $\overline{TV}(p^{\mathrm{d}}, p^{\mathrm{n}})=0.2694$ when the same comparison is made with the description-augmented Choice. This gap indicates that the decision state is not preserved under the $\{0,1\}^K\to\Delta^{K-1}$ projection.

\begin{figure}[t]
\centering
\includegraphics[width=0.99\linewidth]{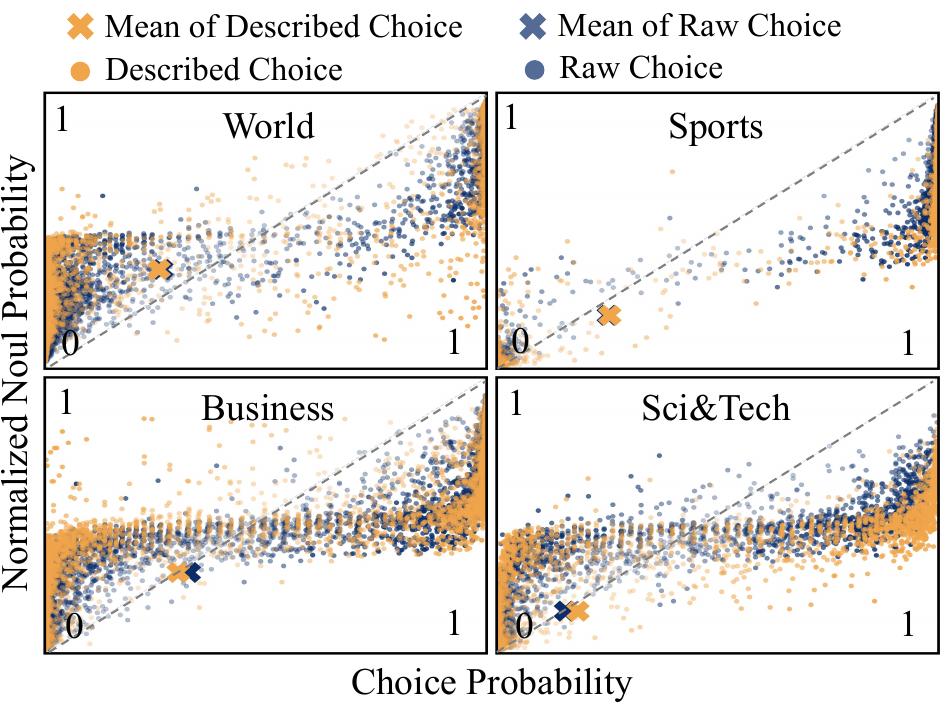}
\vspace{-7pt}
\caption{Axis IV label-description consistency distribution. The four panels correspond to the AG News text categories. The cross symbols (\textbf{$\times$}) denote the mean coordinates of the distributions.}
\label{fig:d4}
\vspace{-16pt}
\end{figure}

To further investigate the violation of \textit{Description Invariance} and \textit{Projection Invariance}, we examine the sample-wise probability distributions across different primitives and choices. As shown in Fig.~\ref{fig:d4}, Choice and Noul probabilities should coincide on the diagonal $y=x$ if both invariances held. Instead, the predictions deviate markedly, exhibiting an asymmetric crossing pattern: when the Choice probability is low, the Noul probability tends to be higher, whereas when the Choice probability is high, Noul becomes more conservative. This indicates that the two primitives interpret the same probability space differently. Furthermore, the cross symbols ($\times$), which denote the distribution means, reveal that for World, Sports, and Sci\&Tech, the mean of the described choice lies farther from the diagonal than that of the raw choice. This demonstrates that adding descriptive text exacerbates cross-primitive inconsistency rather than anchoring the semantics, showing that JevK5's probability simplex is unstable to surface-level perturbations.

\section{Ways to Mitigate Inconsistency}
\begin{figure*}[t]
    \centering
    \includegraphics[width=0.99\linewidth]{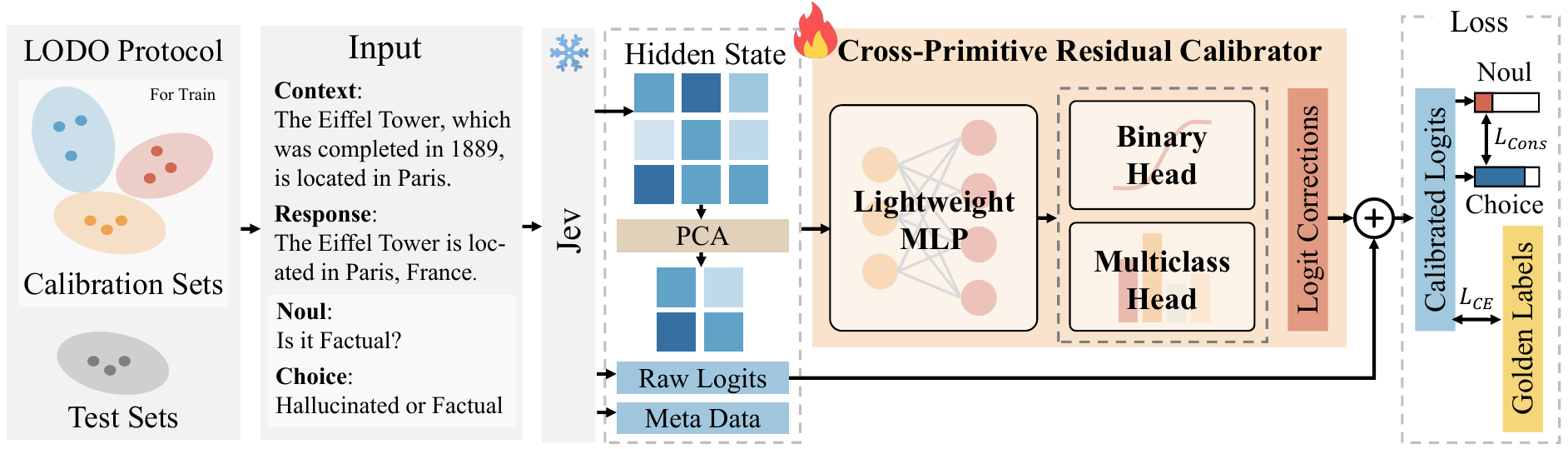}
    \caption{Overview of \textsc{Cprc}.}
    \label{fig:cprc} 
\vspace{-13pt}
\end{figure*}
\textbf{Calibrator Architecture.} 
To mitigate cross-primitive inconsistency and improve Jev's reliability, \textbf{C}ross-\textbf{P}rimitive \textbf{R}esidual \textbf{C}alibrator (\textsc{Cprc}) is proposed. \textsc{Cprc} is a residual calibrator that aligns the predictions of multiple primitives on the same semantic state (\cf Fig.~\ref{fig:cprc}). Keeping the backbone model frozen, we append a lightweight MLP residual network. JevK5 consists of $L$ layers. For each primitive, this calibrator receives the PCA-projected hidden state from the $(L-1)$-th layer, the raw probabilities, and view-level metadata (\eg, predictive entropy). Separate binary and multiclass heads predict logit corrections, which we add directly to the original logits:
\begin{equation}
z_i = z_i^{\rm raw} + \Delta_i, \qquad \Delta_i = \tanh\bigl(f_\theta(x_i)\bigr),
\label{eq:cprc_residual}
\end{equation}
where $z_i^{\rm raw}$ is the raw logit, $x_i$ the concatenated calibrator input, $f_\theta$ is \textsc{Cprc}, and $z_i$ the calibrated logit.

\textbf{Training and Evaluation Protocol.} 
To prevent data leakage, we isolate training and evaluation, selecting PCA dimensions and hyperparameters exclusively on the calibration split before evaluating on the test split. Furthermore, to assess cross-domain generalization, we adopt a leave-one-domain-out (LODO) protocol. We train the calibrator on the combined calibration sets of three axes while holding out the remaining axis for testing. We target Axis IV as the held-out test domain, as it involves more diverse comparisons, making it suitable for evaluating generalization.

\textbf{Training Objective.} We optimize the calibrator using a joint objective:
\begin{equation}
\mathcal L = \mathcal L_{\rm CE} + \alpha\mathcal L_{\rm cons},
\label{eq:loss}
\end{equation}
where $\alpha$ weights the consistency penalty. For binary decisions, $\mathcal L_{\rm CE}$ is the binary cross-entropy
\begin{equation}
\mathcal L_{\rm CE} = -\frac{1}{N}\sum\nolimits_{i=1}^{N}\bigl[y_i\log p_i + (1-y_i)\log(1-p_i)\bigr],
\label{eq:bi_loss}
\end{equation}
and for multiclass decisions, it is the standard cross-entropy
\begin{equation}
\mathcal L_{\rm CE} = -\frac{1}{N}\sum\nolimits_{i=1}^{N}\log p_{i,y_i},
\label{eq:mc_loss}
\end{equation}
where $N$ is the number of calibration samples, $y_i$ is the ground-truth label, $p_i$ is the calibrated probability assigned to the positive class, and $p_{i,y_i}$ is the calibrated probability assigned to the ground-truth class $y_i$.
The consistency penalty $\mathcal L_{\rm cons}$ targets the semantic divergence to each axis: i) the pairwise probability gap for Axis I (\cf Eq.~\ref{eq:ax1}); ii) the $TV$ between Choice and Noul distributions at three granularities for Axis II (\cf Eq.~\ref{eq:ax2}); iii) the probability gap between Choice abstention and Noul answerability for Axis III (\cf Eq.~\ref{eq:ax31}-\ref{eq:ax32}); iv) and the TV between Choice and Noul category-wise predictions for Axis IV (\cf Eq.~\ref{eq:ax41}-\ref{eq:ax42}).

\textbf{Calibration Results.} Beyond measuring the consistency metrics of each axis, we additionally report accuracy.

\textit{In-Domain Evaluation.} Table~\ref{tab:indomain_cons} shows the \textsc{Cprc}'s results on Axis I, II, and III. \textsc{Cprc} mitigates cross-primitive inconsistency across all three axes. For example, on \texttt{JevK5-2B} in Axis I, $G_{n^+,n^-}$ drops from 0.183 to 0.087 ($\textbf{-52.56\%}$) while Choice accuracy rises from 65.03\% to 79.00\% ($\textbf{+13.97\%}$). On \texttt{JevK5-4B} in Axis II, the fine-level divergence $D_F$ drops from 0.633 to 0.355 ($\textbf{-43.95\%}$) while accuracy rises from 95.20\% to 99.65\% ($\textbf{+4.45\%}$). On \texttt{JevK5-2B} in Axis II, $D_M$ drops from 0.481 to 0.130 ($\textbf{-73.03\%}$) while accuracy rises from 77.67\% to 98.48\% ($\textbf{+20.81\%}$).
\begin{table*}[t]
\centering
\small
\setlength{\tabcolsep}{1.2pt}
\renewcommand{\arraystretch}{1.1}
\caption{In-Domain consistency and accuracy. $AG$ denotes the Accuracy Gap (\%).}
\vspace{-2pt}
\label{tab:indomain_cons}
\begin{tabularx}{\textwidth}{@{}l|*{4}{>{\centering\arraybackslash}X}|*{4}{>{\centering\arraybackslash}X}|*{4}{>{\centering\arraybackslash}X}@{}}
\toprule
\multirow{2}{*}{\textbf{Model}} & \multicolumn{4}{c|}{\textbf{Axis I}} & \multicolumn{4}{c|}{\textbf{Axis II}} & \multicolumn{4}{c}{\textbf{Axis III}} \\
\cmidrule(lr){2-5}\cmidrule(lr){6-9}\cmidrule(lr){10-13}
& \textbf{\scalebox{0.91}{$G_{n^+,n^-}\!\!\downarrow$}} & \textbf{\scalebox{0.91}{$G_{n^+,c}\!\downarrow$}} & \textbf{\scalebox{0.91}{$G_{n^-,c}\!\downarrow$}} & Acc
& \textbf{$D_C\downarrow$} & \textbf{$D_M\downarrow$} & \textbf{$D_F\downarrow$} & Acc
& \textbf{$G_A\downarrow$} & \textbf{$K$(\%)$\downarrow$} & $AG\downarrow$ & Acc \\
\midrule
\texttt{JevK5-2B} & 0.1832 & 0.2442 & 0.1106 & 65.03 & 0.2923 & 0.4806 & 0.4867 & 77.67 & \textbf{0.1884} & 36.15 & 8.77 & 77.81 \\
\rowcolor{lightcyan}
\texttt{JevK5-2B} \scalebox{0.78}{(\textsc{Cprc})} & \textbf{0.0869} & \textbf{0.0962} & \textbf{0.0643} & \textbf{79.00} & \textbf{0.0666} & \textbf{0.1296} & \textbf{0.1573} & \textbf{98.48} & 0.2260 & \textbf{18.37} & \textbf{6.92} & \textbf{87.96} \\
\midrule
\texttt{JevK5-4B} & 0.1361 & 0.1132 & 0.0688 & 71.77 & 0.0870 & 0.5330 & 0.6325 & 95.20 & 0.2619 & 26.74 & 11.03 & 85.63 \\
\rowcolor{lightcyan}
\texttt{JevK5-4B} \scalebox{0.78}{(\textsc{Cprc})} & \textbf{0.0577} & \textbf{0.0619} & \textbf{0.0511} & \textbf{82.80} & \textbf{0.0842} & \textbf{0.2815} & \textbf{0.3545} & \textbf{99.65} & \textbf{0.2018} & \textbf{14.84} & \textbf{9.90} & \textbf{93.58} \\
\midrule
\texttt{JevK5-9B} & 0.0861 & 0.0931 & 0.0654 & 71.22 & 0.1292 & 0.3618 & 0.5085 & 93.77 & 0.2103 & 17.27 & 11.51 & 91.79 \\
\rowcolor{lightcyan}
\texttt{JevK5-9B} \scalebox{0.78}{(\textsc{Cprc})} & \textbf{0.0567} & \textbf{0.0633} & \textbf{0.0575} & \textbf{80.98} & \textbf{0.0890} & \textbf{0.1756} & \textbf{0.2534} & \textbf{99.13} & \textbf{0.1802} & \textbf{12.83} & \textbf{9.73} & \textbf{95.67} \\
\bottomrule
\end{tabularx}
\end{table*}

\textit{Out-of-Domain Generalization.} To evaluate whether \textsc{Cprc} learns fundamental corrections rather than memorizing domain-specific statistics, we test it under the LODO protocol, with Axis IV as the unseen target domain. The results show that \textsc{Cprc} generalizes its consistency improvements to this unseen domain (Table~\ref{tab:ood_cons}). For example, on \texttt{JevK5-2B}, $\overline{TV}(p^{\mathrm{r}}, p^{\mathrm{n}})$ drops from 0.439 to 0.182 ($\textbf{-58.57\%}$) while raw-label accuracy rises from 84.90\% to 87.42\% ($\textbf{+2.97\%}$). The same model also reduces $\overline{TV}(p^{\mathrm{d}}, p^{\mathrm{n}})$ from 0.474 to 0.237 ($\textbf{-49.98\%}$) while description-augmented accuracy rises from 86.80\% to 87.90\% ($\textbf{+1.27\%}$). On \texttt{JevK5-9B}, $\overline{TV}(p^{\mathrm{r}}, p^{\mathrm{n}})$ drops from 0.225 to 0.094 ($\textbf{-58.42\%}$) while raw-label accuracy rises from 88.02\% to 89.75\% ($\textbf{+1.96\%}$). Averaged over 6,000 test instances, \textsc{Cprc} introduces only a \textbf{+0.75\%} additional latency overhead, which is negligible.

\begin{table}[htbp]
\centering
\small
\setlength{\tabcolsep}{3pt}
\renewcommand{\arraystretch}{1.1}
\caption{Out-of-Domain consistency.}
\vspace{-5pt}
\label{tab:ood_cons}
\begin{tabularx}{\columnwidth}{@{}l|*{3}{>{\centering\arraybackslash}X}@{}}
\toprule
\multirow{2}{*}{\textbf{Model}} & \multicolumn{3}{c}{\textbf{Consistency ($\overline{TV} \downarrow$)}} \\
\cmidrule(lr){2-4}
& \textbf{$\overline{TV}(p^{\mathrm{r}}, p^{\mathrm{d}})$} & \textbf{$\overline{TV}(p^{\mathrm{r}}, p^{\mathrm{n}})$} & \textbf{$\overline{TV}(p^{\mathrm{d}}, p^{\mathrm{n}})$} \\
\midrule
\texttt{JevK5-2B} & \textbf{0.0825} & 0.4391 & 0.4736 \\
\rowcolor{lightcyan}
\texttt{JevK5-2B} \scalebox{0.85}{\textsc{(Cprc)}} & 0.0988 & \textbf{0.1819} & \textbf{0.2369} \\
\midrule
\texttt{JevK5-4B} & 0.0907 & 0.2407 & 0.2696 \\
\rowcolor{lightcyan}
\texttt{JevK5-4B} \scalebox{0.85}{\textsc{(Cprc)}} & \textbf{0.0663} & \textbf{0.1719} & \textbf{0.2083} \\
\midrule
\texttt{JevK5-9B} & \textbf{0.0447} & 0.2251 & 0.2477 \\
\rowcolor{lightcyan}
\texttt{JevK5-9B} \scalebox{0.85}{\textsc{(Cprc)}} & 0.0513 & \textbf{0.0936} & \textbf{0.1262} \\
\bottomrule
\end{tabularx}
\end{table}

\begin{table}[htbp]
\centering
\small
\setlength{\tabcolsep}{3pt}
\renewcommand{\arraystretch}{1.1}
\caption{Out-of-Domain accuracy.}
\vspace{-5pt}
\label{tab:ood_acc}
\begin{tabularx}{\columnwidth}{@{}l|*{2}{>{\centering\arraybackslash}X}@{}}
\toprule
\textbf{Model} & \textbf{Raw Labels} & \textbf{Desc Labels} \\
\midrule
\texttt{JevK5-2B} & 84.90 & 86.80 \\
\rowcolor{lightcyan}
\texttt{JevK5-2B} \scalebox{0.85}{\textsc{(Cprc)}} & \textbf{87.42} & \textbf{87.90} \\
\midrule
\texttt{JevK5-4B} & 84.08 & 86.97 \\
\rowcolor{lightcyan}
\texttt{JevK5-4B} \scalebox{0.85}{\textsc{(Cprc)}} & \textbf{84.12} & \textbf{87.05} \\
\midrule
\texttt{JevK5-9B} & 88.02 & 89.03 \\
\rowcolor{lightcyan}
\texttt{JevK5-9B} \scalebox{0.85}{\textsc{(Cprc)}} & \textbf{89.75} & \textbf{90.07} \\
\bottomrule
\end{tabularx}
\vspace{-12pt}
\end{table}

\section{Related Work}
\noindent\textbf{System-One Decision Model} reduces the cost and improves the speed of using LLMs as decision models through System One-style single-pass inference. Existing Jev-related methods are broadly categorized into three streams. The first category is Empirical Evaluation and Capability Distillation. These methods benchmark system limits or distill decision priors to optimize inference efficiency, \eg, probing zero-shot reasoning boundaries across diverse tasks~\citep{deusser2026evaluating,yang2026can,li2026specialized,porcedda2026jev} and building Jev-style decision models via direct logit readout or low-rank adaptation~\citep{li2026llm2jev,geuter2026releasing,li2026jev,tang2026typed,weng2026jevtypeddecisionsforward,ma2026fastjev}.
The second category is Domain-Specific Adaptation. These methods fine-tune decision layers to target specialized scenarios, \eg, automating clinical and medical evaluations~\citep{gershon2026decisionlanguagemodelsaffordable,madridgarcía2026jevmedicinebenchmarkevaluation} and integrating multilingual or normative knowledge~\citep{wang2026chinese,barbosa2026calibrated,li2026dknown}.
The third category is Adversarial Vulnerability and Safety Assessment. These methods perturb inputs or exploit contextual shifts to expose decision fragility, \eg, executing black-box adversarial edits and prompt injections~\citep{li2026same,hu2026jevadvbench,wu2026decision} and demonstrating susceptibility to natural context flips~\citep{xu2026jevout}, alongside deploying them as zero-shot safety detectors~\citep{guo2026just}.

Despite delineating external performance boundaries, these methods only evaluate input-output pairs without verifying internal coherence, introducing reliability risks when semantically equivalent queries yield contradictory decisions. Unlike these methods, we construct \textsc{PrimaCons}, which quantifies Jev's internal consistency along four axes, and we further propose \textsc{Cprc}, a residual calibrator that mitigates the observed inconsistency with the frozen Jev model.

\noindent\textbf{Consistency of LLMs} refers to the invariance of their outputs under semantically equivalent inputs. Existing studies are broadly categorized into three streams.
The first category is Contextual and Semantic Perturbation Robustness. These methods inject surface-level noise to expose output instability, \eg, mutating prompts while preserving logical semantics to detect behavioral drift~\citep{de2025metamorphic,shi2025sproc,li2025meaning,cox2025mapping,fu2025same,serbout2025quirx} and evaluating reasoning degradation under adversarial inputs, irrelevant distractors, or counterfactual evidence~\citep{zhang2026llm,yang2025llm,mo2026faithfulness,lee2026lost,lin2025breaking,zhao2025large,pandia2021sorting,zhang2025cchall}.
The second category is Internal Representation and Quantization Sensitivity. These methods perturb weights and activations to analyze collapse during low-bit compression, \eg, isolating activation spikes and positional encoding artifacts~\citep{gong2024makes,chen2026massive,qiao2025rethinking,kim2026neural}, tracking how quantization shrinks decision margins~\citep{sathyanarayanan2026perplexity,singh2026saying,li2026quantizationmeetsreasoningexploring,li2026quantizationmeetsreasoningexploring,lotfi2026quantized}, and probing rescaling invariance, feedback exposure, and value-magnitude sensitivity in low-bit rounding~\citep{li2026quantizing,wu2026decisions,arcuschin2026chainofthought,li2026reasoning}.
The third category is Intermediate Reasoning Trace Faithfulness. These methods audit multi-step generation pipelines to detect logical disconnects, \eg, exposing rationales that fabricate evidence for predetermined answers~\citep{yang2025llm,bao2026ratio}, optimizing intermediate token probabilities via consistency decoding or confidence-weighted aggregation, probing commitment misalignment~\citep{li2026reasoning}, answer-reasoning disconnect~\citep{lotfi2026quantized,de2025metamorphic,liu2026let}, and semantic drift~\citep{cheng2025integrative,taubenfeld2025confidence,liu2026sinktrack}.

Our work builds on the unique architecture of decision models, in which the context is decoupled from the querying primitive. It raises a question that single-primitive architecture cannot capture: whether the same state yields a consistent decision across primitives, which directly reflects the model's decision reliability. We quantify this cross-primitive consistency, grounding each axis in a distinct theoretical requirement (binary isomorphism, taxonomic coherence, procedure invariance, description invariance, and projection invariance) and propose \textsc{Cprc}, a calibrator that aligns the probability distributions across primitives.
\section{Conclusion}
\label{sec:conclusion}

System-One decision models bypass reasoning to accelerate inference. Their reliability is therefore hard to audit, yet it can still be assessed through consistency. Cross-primitive consistency offers one such lens. Therefore, we introduce \textsc{PrimaCons} to quantify it, revealing that current decision models violate different invariants across complementary, hierarchical, abstention, and label-description dimensions. To mitigate this inconsistency, we propose \textsc{Cprc}, which appends a lightweight residual network and optimizes the network while keeping the backbone frozen via a joint consistency objective. Validated under an LODO protocol, \textsc{Cprc} curtails inconsistency and boosts accuracy across unseen domains. We hope this work offers a lens for assessing decision-model reliability and that the proposed calibrator helps make such models deployable in high-risk scenarios.

{
    \small
    \bibliographystyle{ieeenat_fullname}
    \bibliography{main}
}


\end{document}